# Autonomous Driving Strategies at Intersections: Scenarios, State-of-the-Art, and Future Outlooks

Lianzhen Wei, Zirui Li, Jianwei Gong, Cheng Gong, Jiachen Li

*Abstract*— Due to the complex and dynamic character of intersection scenarios, the autonomous driving strategy at intersections has been a difficult problem and a hot point in the research of intelligent transportation systems in recent years. This paper gives a brief summary of state-of-the-art autonomous driving strategies at intersections. Firstly, we enumerate and analyze common types of intersection scenarios, corresponding simulation platforms, as well as related datasets. Secondly, by reviewing previous studies, we have summarized characteristics of existing autonomous driving strategies and classified them into several categories. Finally, we point out problems of the existing autonomous driving strategies and put forward several valuable research outlooks.

## I. INTRODUCTION

As a typical traffic scenario, the intersection plays an important role but also brings serious problems, which have become the focus of many researchers. According to the U.S. National Highway Traffic Safety Administration's fatality analysis report, more than one-fourth of fatal crashes in the U.S. occur at or are related to the intersection, and about 50% of these occur at uncontrolled intersections [1]. In addition to frequent safety-related accidents, intersection congestion often occurs, which brings huge economic loss and environmental pollution around the world.

A great many efforts have been done in different aspects to tackle the above problems. According to the strategy implemented, solutions can be roughly classified into two categories: (a) research about the traffic structure; (b) research about the autonomous vehicle [2]. The first kind of research focuses on the topological characteristic and the signal control method at intersections. Adopting different intersection topologies in different situations is proved to be an effective measure to dredge traffic flow and ensure interaction safety [3]. Also, some researchers focus on optimizing the control of traffic lights to improve the efficiency of the intersection. Lian et al. have proposed two adaptive signal control algorithms which are effective for easing traffic congestion and achieve adaptive signal control objectives using real-time traffic information [4]. Soheil et al. have proposed an adaptive traffic signal controller based on reinforcement learning, which can receive non-preset high-dimensional sensory information and perform self-learning to minimize the delay at intersections [5]. Although these signal control strategies can partially improve the traffic flow if all approaches to the intersection are not equally congested, they cannot eliminate the stop delay of vehicles at intersections regardless of the traffic volume [6]. With the development of intelligent network communication as well as autonomous driving algorithm, autonomous driving technology has become an important way to solve the above problems. Some researchers focus on the autonomous driving strategy at intersections to improve traffic efficiency and reduce pollution to a greater extent.

With the accumulation of research on intersection driving strategies, it is time to classify and organize them. Throughout the past, we can see several overviews about intersections or about driving strategies, but the two are not well integrated. For instance, overviews on the intersection monitoring or the scheduling are presented in [7-9]. However, these studies rarely analyze and solve intersection problems from the perspective of autonomous driving vehicles. There are also some reviews on the decision-making for autonomous driving vehicles in [2, 10]. However, these studies only focus on autonomous driving vehicles and have not made an analysis for the specific scenario of intersections. Different from the above reviews, we narrow the scope of our research and focus on the autonomous driving strategy at intersections. In addition to summarizing autonomous driving strategies of the intersection scenario, this paper also contains the following features: 1) The categories of existing intersection scenarios are divided, and the characteristics of intersection scenarios are introduced. 2) The simulation platforms and datasets related to the intersection have been summarized. 3) The inadequacy of the existing autonomous driving strategies at intersections is analyzed, and future outlooks are forecasted.

The rest of this paper is organized as follows. Section II presents a detailed introduction of the outline, including different types and characteristics of intersections, simulation platforms, as well as related datasets. Section III summarizes state-of-the-art autonomous driving strategies at intersections into several categories and analyzes each type of strategy. Section IV points out the inadequacy of existing autonomous driving strategies and puts forward future outlooks. Finally, we make a general conclusion about the full text in section V.

## II. THE OUTLINE OF INTERSECTIONS

In order to show a comprehensive and detailed description of intersection scenarios, we have summarized an outline in this section. It includes the common types and characteristics of intersections, corresponding simulation platforms, as well as related open source datasets.

*This research was supported by the National Natural Science Foundation of China under grant number U19A2083.

Lianzhen Wei, Zirui Li, Jianwei Gong, Cheng Gong are from the School of Mechanical Engineering, Beijing Institute of Technology, Beijing, China. (E-mails: 3120200396@bit.edu.cn; 3120195255@bit.edu.cn; gongjianwei @bit.edu.cn; chenggong@bit.edu.cn). Jiachen Li is with University of California, Berkeley, USA (E-mail: jiachen_li@berkeley.edu).

Zirui Li is also from the Department of Transport and Planning, Faculty of Civil Engineering and Geosciences, Delft University of Technology, Stevinweg 1, 2628 CN Delft, The Netherlands.

(Corresponding author: Jianwei Gong and Zirui Li)

*A. Types and Characteristics of Intersections*

Autonomous driving at intersections is difficult mainly due to complex traffic conditions. Firstly, there are different types of participants, such as vehicles, motorcycles, bicycles, and pedestrians. Their movements are sometimes irregular and unpredictable. Secondly, The intersection structure itself is very complicated in terms of topology. Here, we can classify intersections at grade into several main categories based on their topology: crossroad, X-intersection, Y-intersection, T-intersection, roundabout, misaligned intersection, ramp merge, and deformed intersection. These intersections are illustrated in Figure 1.

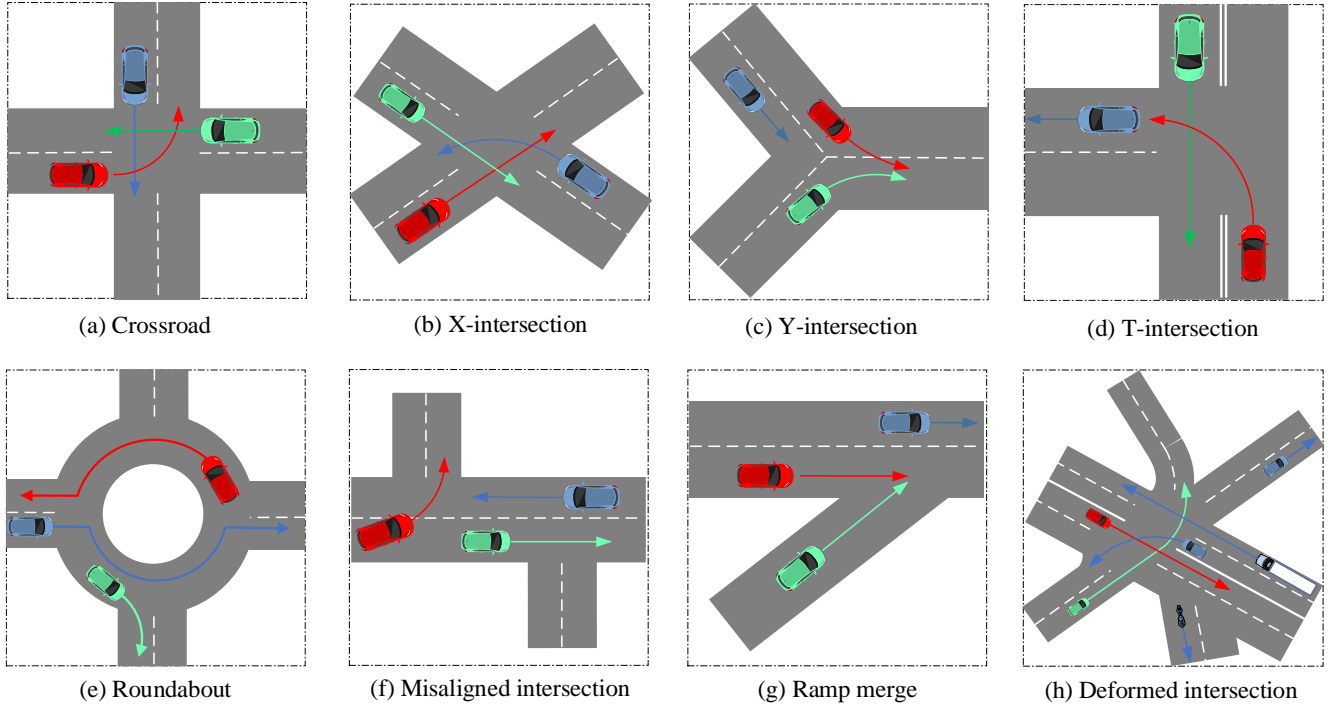

(a) Crossroad  (b) X-intersection  (c) Y-intersection  (d) T-intersection

(e) Roundabout  (f) Misaligned intersection  (g) Ramp merge  (h) Deformed intersection

Figure 1. Different types of intersections

The main difference between intersections and other road structures is that there are more conflict points at intersections [11]. So many conflict points lead to a greater probability of collisions between vehicles. In particular, the left turn of vehicles is more likely to collide than the straight or the right turn [12]. To improve the safety and efficiency of vehicles passing through the intersection, many researchers have made contributions. Among them, there are many simulations that can be applied to intersection algorithm testing, valuable datasets, as well as a variety of driving strategies. From the perspective of driving strategy, the current research pays more attention to the autonomous driving algorithm of the isolated intersection, such as [13, 14]. In fact, vehicles in the road network will pass through multiple intersections in turn, and vehicles at multiple intersections are relevant in the long term. Considering that the optimization of a single intersection will finitely promote the overall traffic flow, Some scholars have focused on the global collaborative optimization of multiple intersections such as [15, 16].

*B. Simulation Platforms*

To shorten the time cost, reduce the economic cost and ensure safety, the training and testing of autonomous driving strategy usually start from simulation platforms. Here, we have summarized the commonly used simulation platforms for autonomous driving strategies at intersections in Table I. Some of them are commercial, while the others are open source and can be used for free.

Different simulation platforms have different focuses and we divide them into three sub-categories. Well, the barriers between simulation platforms are gradually disappearing. To make the simulation results closer to the real effect, co-simulation is becoming more and more common. Also, the combination of real datasets and real maps in the simulation platform is underway.

*1) Special Scene Simulation Platforms*

Some simulation platforms like FLUIDS focus on the simulation of the special scene including intersections. These simulation platforms have a small scope of application, but have strong pertinence and are very suitable for intersection research.

*2) Traffic Flow Simulation Platforms*

Some simulation platforms like PTV Vissim and SUMO focus on the traffic flow modeling of intelligent transportation systems. These platforms can be combined with real maps to provide more complete traffic road information for traffic flow control.

*3) Autonomous Driving Vehicle Simulation Platforms*

Some simulation platforms like CARLA, PreScan, and Carsim focus on the simulation of autonomous driving vehicles. They provide predetermined maps, vehicle models, as well as sensor models. These simulation platforms usually have rendering engines like Unreal Engine or Unity to simulate the external environment more realistically.

TABLE I. CORRESPONDING SIMULATION PLATFORMS

| Simulator | Provider | Supported Systems | Type | URL | Reference |
|---|---|---|---|---|---|
| FLUIDS | Berkeley AUTOLAB | Linux | Open Source | *https://urban-driving-simulator.readthedocs.io/* | [17] |
| PTV Vissim | PTV Group | Windows, Linux | Commercial | *https://www.ptvgroup.com/en/solutions/products/ptv-vissim/* | [18] |
| SUMO | Eclipse Foundation | Windows, Linux, macOS | Open Source | *https://www.eclipse.org/sumo/* | [19] |
| PreScan | TASS International | Windows | Commercial | *https://tass.plm.automation.siemens.com/prescan/* | [20] |
| CARLA | CVC | Windows, Linux | Open Source | *https://carla.org/* | [21] |
| Carsim | Mechanical Simulation | Windows, Linux | Commercial | *https://www.carsim.com/products/carsim/* | [22] |
| SUMMIT | AdaCompNUS | Linux | Open Source | *https://adacompnus.github.io/summit-docs/* | [23] |
| LGSVL | LG Electronics | Windows, Linux | Open Source | *https://www.svlsimulator.com/* | [24] |
| Apollo | Baidu | Windows, Linux | Open Source | *https://apollo.auto/gamesim.html* | [25] |
| AirSim | Microsoft | Windows, Linux, macOS | Open Source | *https://microsoft.github.io/AirSim/* | [26] |
| VTD | Hexagon | Linux | Commercial | *https://vires.mscsoftware.com/* | [27] |
| 51Sim-One | 51WORLD | Windows, Linux | Commercial | *https://www.51aes.com/values/simulation* | [28] |
| ASM Traffic | dSPACE | Windows | Commercial | *https://www.dspace.com/en/pub/home/medien/product_info/prodinf_asm_traffic.cfm* | [29] |
| DYNA4 | VECTOR | Windows | Commercial | *https://www.vector.com/int/en/products/products-a-z/software/dyna4/* | [30] |
| SCANeR Studio | AVSimulation | Windwos | Commercial | *https://www.avsimulation.com/scaner-studio/* | [31] |
| PanoSim | PanoSim Technologies | Windwos | Commercial | *http://www.panosim.com/* | [32] |
| MATLAB | MathWorks | Windwos | Commercial | *https://www.mathworks.com/products/matlab.html* | [33] |

*C. Related Datasets*

Datasets can be used for driving strategy training or verification, which is of great significance for the development of intelligent transportation systems and the progress of autonomous driving technologies. Over the past decade, a number of datasets for autonomous driving have been made public by numerous institutions around the world.

According to the purpose of datasets, they can be divided into object recognition, object tracking, road/lane detection, semantic segmentation, behavior analysis, end-to-end learning, etc. Junyao Guo et al. have sorted out existing autonomous driving datasets and classified them according to autonomous driving tasks [34]. They have summarized a lot of datasets, but these datasets are not specifically for intersection scenarios. Here, we have combined several reviews of datasets and newly released datasets to specifically list some of those that are highly related to intersections. To facilitate the acquisition of datasets, we have given their links. In addition, we also shared a dataset website which is maintained by GRAVITI: *https://gas.graviti.cn/open-datasets*.

Among them, some datasets are specifically for scenarios of intersections like ACFR. There are also some datasets related to traffic sign recognition like Tsinghua-Tencent 100K. These datasets are helpful for training autonomous vehicles to recognize traffic signs at intersections. In addition, some newly released datasets can support the completion of rich training or testing tasks, such as object recognition and tracking, semantic understanding, lane/road detection, pedestrian prediction, etc. These datasets are also of great significance in intersection research.

TABLE II. RELATED OPEN SOURCE DATASETS

| Name | Provider | Data Contents | URL | Reference |
|---|---|---|---|---|
| ACFR | The University of Sydney | Vehicle trajectories at five roundabouts | *http://its.acfr.usyd.edu.au/datasets/five-roundabouts-dataset/* | [35] |
| MTID | BEUMER Group | Intersection scenario data | *https://vap.aau.dk/dataset/* | [36] |
| Tsinghua-Tencent 100K | Tsinghua; Tencent | Traffic sign recognition | *https://cg.cs.tsinghua.edu.cn/traffic-sign/* | [37] |
| MTSD | Mapillary | Traffic sign recognition | *https://www.mapillary.com/dataset/trafficsign* | [38] |
| ApolloScape | Baidu | Trajectory prediction; 3D lidar object detection and tracking; Stereo estimation; Lanemark segmentation; Online self-localization; 3D car instance understanding | *http://apolloscape.auto/* | [39] |
| Mapillary Vistas Dataset | Mapillary | Visual road-scene understanding | *http://eval-vistas.mapillary.com/* | [40] |
| KITTI | KIT; TTIC | Odometry; Object detection; Tracking benchmarks; Road benchmark | *http://www.cvlibs.net/datasets/kitti/* | [41] |
| Cityscapes | Daimler AG R&D; Max Planck Institute; TU Darmstadt | Semantic understanding of urban street scenes | *https://www.cityscapes-dataset.com/* | [42] |
| JAAD | Yorku EECS | Object detection and tracking; Behavior analysis | *http://data.nvision2.eecs.yorku.ca/JAAD_dataset/* | [43] |
| BDD100K | The University of Berkeley | Object detection and tracking; Semantic segmentation; Lane/road detection | *https://bdd-data.berkeley.edu/* | [44] |
| H3D | Honda Research Institute | Object detection and tracking | *http://usa.honda-ri.com/H3D* | [45] |
| nuScenes | Motional | Object detection and tracking; Semantic segmentation; Object prediction | *https://www.nuscenes.org/* | [46] |

### III. REVIEW OF DRIVING STRATEGIES AT INTERSECTIONS

In order to improve traffic efficiency and reduce the pollution caused by vehicle delay to a greater extent, numerous researchers focus on the autonomous driving strategy at intersections. In this chapter, we make a concise summary of state-of-the-art autonomous driving strategies at intersections. Firstly, the autonomous driving strategy is divided into the cooperative driving strategy and the individual driving strategy. The cooperative driving strategy can be divided into the centralized driving strategy and the distributed driving strategy while the individual driving strategy can be divided into the classical driving strategy and the learning-based driving strategy. Each strategy includes many specific methods, as shown in Figure 2.

#### A. The Cooperative Driving Strategy

The core idea of cooperative driving is to construct and execute a global optimal sequence for vehicles [2]. According to the architecture of computing and control, we divide the cooperative driving strategy into two types: the centralized driving strategy and the distributed driving strategy.

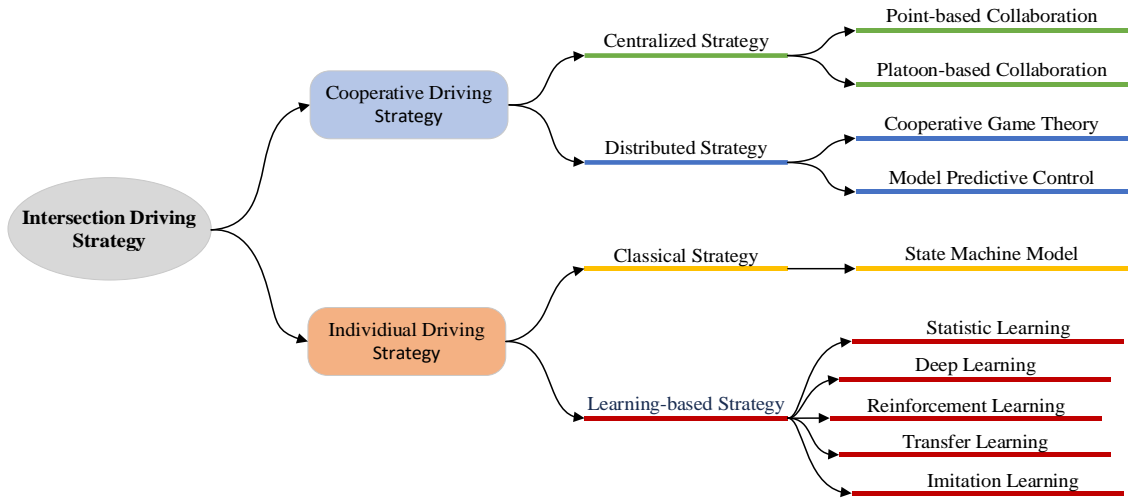

Figure 2. State-of-the-art autonomous driving strategies at intersections

*1) The Centralized Driving Strategy*

The centralized driving strategy mainly relies on vehicle-to-infrastructure (V2I) technology to achieve two-way communication between vehicles and road infrastructures. Obviously, the centralized driving strategy requires the intersection coordination unit (ICU) to schedule vehicles in a certain range around the intersection [6]. In the process of interaction, vehicles are supposed to send their information, such as location, speed, driving intention, etc. to the ICU. Then, the ICU carries out the centralized control through a specific scheduling algorithm and sends the high-level control information to vehicles. Vehicles can receive the high-level command and carry out the low-level control individually. The existing centralized cooperative driving strategy can also be divided into two the point-based collaboration method and the platoon-based collaboration method. The former treats each vehicle as a specific point for priority division, and the latter first groups the vehicles, and then coordinates the passing order of different groups.

The point-based collaboration means that the ICU assigns different priorities to individual vehicles near the intersection in some way. The vehicle with higher priority passes first, and the vehicle with lower priority passes later. If the time difference for two vehicles reaching the conflict point were less than the threshold, the longitudinal speeds of the two vehicles would be adjusted to ensure safety. There are many ways to divide vehicle priority like the rule-based method, the search-based method, the optimization-based method, and so on. In [14], a new method based on heuristic rules and Monte Carlo tree search is proposed, which can obtain a suboptimal vehicle sequence in a short time. In [47], a priority-based traffic architecture is proposed. The ICU gives priority to all intersection participants based on specific rules. In [48], the multi-vehicle traffic problem is modeled as a multi-objective optimization problem. The objective function includes delay time, fuel consumption, comfort, and other indicators. The constraint condition is usually the allowable speed range of vehicles and the time difference reaching the conflict point. The acceleration or deceleration of the vehicle can be used as the control quantity. In [49], two optimal allocation methods of vehicle traffic sequence are studied, which are dynamic programming and Petri Nets model.

The platoon-based collaboration method groups vehicles into several platoons before they arrive in the vicinity of the intersection. vehicles pass through the intersection safely and efficiently according to the set flow queue mode. In [50], vehicles arriving at the intersection are divided into discrete groups, and the continuous traffic flow problem is simplified to a multi-vehicle cooperation problem with limited vehicles. In [51], a cooperative framework is proposed. The scheduling algorithm frequently plans strategies according to the real traffic situation to guide vehicle rearrangement. In [52], a hierarchical centralized control strategy is proposed. Firstly, vehicles are listed as the standard traffic flow queue before they arrive at the intersection. Secondly, the ICU uses the optimal control method to deal with the traffic queue. It is worth mentioning that the optimal control operation can be carried out offline in advance, while the online matching is carried out according to the actual situation. In [53], a platoon-based cooperative strategy for mixed autonomous and human-driven vehicles at intersections is proposed. It assumes that an autonomous vehicle is leading, and other hybrid vehicles follow behind. This orderly arrangement of traffic can reduce energy consumption. In summary, the centralized driving strategy can achieve global optimization [14]. However, the amount of calculation will increase significantly with the increase of vehicle numbers, which brings a huge challenge to solution calculation.

*2) The Distributed Driving Strategy*

The distributed driving strategy uses vehicle-to-vehicle (V2V) technology to communicate with each other between vehicles to avoid collisions. Compared with the centralized driving strategy, it has the following advantages. 1) No ICU is required, so the economic cost of transportation equipment construction can be reduced. 2) The amount of computation is dispersed, so it has better real-time and scalability. However, the distributed driving strategy can only obtain a suboptimal solution rather than the global optimal solution [14]. Under this strategy, vehicles will pass through intersections on a first-come-first-served basis, and some adjustments may be made based on the specific control method [54].

Cooperative game theory is a suitable tool for modeling strategic interaction between autonomous driving vehicles and it has been exploited for vehicle interactions at intersections by a lot of researchers. In [1], a multi-vehicle interaction strategy based on the leader-follower game is proposed. According to specific rules, the ego-vehicle determines the leader-follower relationship with other vehicles to make the corresponding operation. In [55], a vehicle interactive reasoning model based on fuzzy logic is established. A conflict decision model of intelligent vehicles based on the game theory is proposed. In [56], an autonomous driving strategy based on the cooperative game theory is proposed to avoid vehicle collisions at intersections. In [57], a cooperative decision-making method based on the coalitional game theory is proposed to solve the multi-lane merging problem for autonomous driving vehicles.

Model predictive control is a special optimal control method and has been widely studied in recent years. It calculates the future multi-step control quantity, but the autonomous vehicle only executes the latest one in each control cycle. The model predictive control will be carried out again in the next cycle. In this way, the autonomous vehicle can deal with emergencies immediately. In [58], a distributed MPC algorithm is used to realize the multi-vehicle cooperative traffic, which improves the traffic efficiency of the intersection. In [13], autonomous driving at intersections is realized according to the first-come-first-served traffic strategy and combined with virtual platooning control. In [59], the intersection control problem without signal is modeled as a distributed optimal control problem.

*B. The Individual Driving Strategy*

Although the cooperative driving strategy has considerable advantages, there are also some problems. 1) The cooperative driving strategy relies too much on wireless communication. Once there is a large delay or error in communication, it will have a catastrophic impact on the safety of intersection traffic. 2) Most participants will also be human-driving vehicles in the near future and it is difficult for autonomous driving vehicles to establish robust communication with human drivers. Thus, the research on individual driving strategies is also valuable. The individual driving strategy means that the autonomous vehicle can take appropriate decisions independently without using the V2I or V2V technology, mainly depends on itself.

*1) The Classical Strategy*

The representative classical strategy is the state machine model. Finite State Machine (FSM) divides the state of vehicles into a limited number of categories. When the external scenario changes, the autonomous driving vehicle takes corresponding measures according to the predefined state change rules. Hierarchical State Machine (HSM) is an improvement on the basis of FSM. It divides original parallel classes into several layers to improve maintainability and expandability. Thanks to its good stability and easy operation, the state machine model has been widely used. For instance, many teams in the DARPA challenge have adopted the HSM as their decision-making method [60, 61]. In [12], the FSM is considered to use at the T-intersection. However, the classical method is more suitable for simple scenarios rather than complex dynamic scenarios, because the artificially defined rules cannot adapt to all situations [62].

*2) The Learning-based Strategy*

The learning-based strategy has many detailed categories including statistic learning, deep learning, reinforcement learning, imitation learning, transfer learning, etc. Compared with the classical strategy, the learning-based strategy uses a large amount of data for training to obtain decision-making ability in a complex dynamic environment. In [63], Hidden Markov Model (HMM) is used to predict the intention of other vehicles at intersections, and the Partially Observable

Markov Decision Process (POMDP) method is used to make decisions. In [64], an intention inference method of vehicles based on the Support Vector Machine (SVM) has been proposed. In [65], reinforcement learning is proposed to achieve autonomous driving at intersections. In [66], the transfer learning method is used to classify intersections. In [67], an autonomous driving simulation experiment is carried out through conditional imitation learning. In essence, the learning-based strategy is more suitable for complex dynamic scenarios. However, the learning-based strategy has high training costs and is difficult to realize semantic interpretation. Recently, some researchers focus on interpretable learning algorithms and lifelong learning algorithms to solve the above shortcomings [68, 69].

## IV. Problems and Future Outlooks

In this section, we focus on the problems of the current autonomous driving strategies as well as the future research outlooks at intersections.

### A. Problems

Although some useful methods of autonomous driving at intersections have been brought up or tested, there are still some problems to be addressed. Firstly, when the number of vehicles increases, how to ensure the real-time and robustness of the algorithm is a great challenge. Most of the existing autonomous driving strategies at intersections rely on V2V or V2I technology. In this case, it is of great significance to ensure the algorithm is still safe under the condition of certain sensing noise and communication delay. In addition, most of the above strategies are based on scenarios that all the participants are autonomous driving vehicles, ignoring highly complex participants such as human-driving vehicles and pedestrians. How to ensure smooth and safe traffic in such a complex situation with many participants is really difficult.

### B. Future Outlooks

For now, the autonomous driving vehicle is not as flexible as the vehicle driven by an experienced human driver when dealing with a complex dynamic environment. For example, autonomous vehicles may be in deadlocks or collisions because of a failure to understand the interactions [70]. To improve the level of autonomous driving at intersections, we have listed some future outlooks worthy of study here.

*1) The Cognition-based Driving Strategy*

The main reason why autonomous vehicles cannot perform well in complex dynamic scenes is that they have not formed a good knowledge and understanding of the environment. The incomprehension of the environment contains two important aspects, one is the incomprehension of the input information, and the other is the incomprehension of the decision-making behavior. More advanced autonomous driving strategies should incorporate stronger cognitive abilities. The cognitive ability includes the understanding of the environmental semantics, the understanding of driving behavior, the interpretability of algorithms, and so on.

*2) The Transfer-based Driving Strategy*

Sometimes, autonomous driving vehicles may deadlock or collide in a complex dynamic environment. The former is often because the algorithm is too conservative, while the latter is often because the algorithm is too aggressive. However, an experienced human driver can accurately analyze the intentions of other participants and flexibly make appropriate actions in this situation. In recent years, some researchers have paid attention to driver behavior transfer or imitating driver's actions based on the learning-based approach [71-74]. This is a promising research direction and needs further research.

*3) The Interaction-based Driving Strategy*

As we know, the reason why the intersection is more challenging for autonomous driving vehicles than other scenarios is that it has greater dynamic characteristics and uncertainty. In addition to other autonomous driving vehicles, there are also participants such as human-driving vehicles and pedestrians near the intersection, and their actions are sometimes irregular and even violate traffic laws. To better capture the intention of surrounding dynamic participants, it is necessary to establish interaction with them. The prediction of dynamic participants around the autonomous vehicle has become the focus of some researchers [75-80]. In addition, vehicle-to-pedestrian (V2P) is also a promising way [81].

*4) The Fusion-based Driving Strategy*

Each method has its own shortcomings, and the fusion of methods is an effective method to improve the robustness of the system. For example, the learning-based driving strategy can be combined with the rule-based driving strategy, and the centralized driving strategy can be combined with the distributed driving strategy. Also, multiple methods in the same driving strategy can also be combined. In the perception level of the autonomous driving framework, multi-sensor fusion has been widely used [82]. The integration of multiple driving strategies is conducive to complement each other's deficiencies to improve overall performance. So, why not apply this integration strategy to the decision-making and planning levels as well?

## V. Conclusion

As an important traffic element, the intersection plays a crucial role in the efficiency, safety, and energy consumption of the whole traffic. In this article, we focus on autonomous driving at intersections and make an overview. In summary, the intersection is still a challenging scenario for autonomous driving. With the continuous maturity of technologies such as cognition, prediction, and networking, we believe that traffic at intersections will be greatly improved.